\documentclass[final,5p,times,twocolumn,fleqn]{elsarticle}

\usepackage{amssymb}
\usepackage{amsthm}
\biboptions{sort&compress}

\usepackage{newtxmath}

\usepackage{graphicx}
\usepackage{float} 
\usepackage{tabularx}   
\usepackage{booktabs}   
\usepackage{newtxtext}
\usepackage[numbers]{natbib} 
\usepackage[hidelinks]{hyperref}
\usepackage[switch]{lineno}
\usepackage{pdflscape}  
\usepackage{tabularx}   
\usepackage{multirow}
\usepackage{adjustbox}
\usepackage{geometry}
\usepackage{multirow}
\usepackage{tabularx}
\usepackage{graphicx}

\journal{Information Fusion}

\begin{document}

\begin{frontmatter}







\title{HydraNet: Momentum-Driven State Space Duality for Multi-Granularity Tennis Tournaments Analysis}






\author[label1,label4,label6]{Ruijie Li}
\author[label2,label6]{Xiang Zhao}
\author[label1,label3,label5]{Qiao Ning\corref{cor1}}
\author[label1,label4,label6]{Shikai Guo}

\cortext[cor1]{Corresponding author}

\affiliation[label1]{organization={School of Information Science And Technology},
            addressline={Dalian Maritime University},
            city={Dalian},
            postcode={116026},
            country={China}}
\affiliation[label2]{organization={College of Marine Electrical Engineering},
            addressline={Dalian Maritime University},
            city={Dalian},
            postcode={116026},
            country={China}}
\affiliation[label3]{organization={School of Artificial Intelligence and Computer Science},
            addressline={Jiangnan University},
            city={Wuxi},
            postcode={214122},
            country={China}}
\affiliation[label4]{organization={DUT Artificial Intelligence Institute},
            city={Dalian},
            country={China}}
\affiliation[label5]{organization={Key Laboratory of Symbolic Computation and Knowledge Engineering of Ministry of Education},
            addressline={Jilin University},
            city={Changchun},
            postcode={130012},
            country={China}}
\affiliation[label6]{organization={Dalian Key Laboratory of Artificial Intelligence},
            city={Dalian},
            country={China}}




\begin{abstract}
In tennis tournaments, momentum, a critical yet elusive phenomenon, reflects the dynamic shifts in performance of athletes that can decisively influence match outcomes. 
Despite its significance, momentum in terms of effective modeling and multi-granularity analysis across points, games, sets, and matches in tennis tournaments remains underexplored. 
In this study, we define a novel Momentum Score (MS) metric to quantify a player's momentum level in multi-granularity tennis tournaments, and design HydraNet, a momentum-driven state-space duality-based framework, 
to model MS by integrating thirty-two heterogeneous dimensions of athletes performance in serve, return, psychology and fatigue.
HydraNet integrates a Hydra module, which builds upon a state-space duality (SSD) framework, capturing explicit momentum with a sliding-window mechanism and implicit momentum through cross-game state propagation. It also introduces a novel Versus Learning method to better enhance the adversarial nature of momentum between the two athletes at a macro level, along with a Collaborative-Adversarial Attention Mechanism (CAAM) for capturing and integrating intra-player and inter-player dynamic momentum at a micro level.
Additionally, we construct a million-level tennis cross-tournament dataset spanning from 2012-2023 Wimbledon and 2013-2023 US Open, and validate the multi-granularity modeling capability of HydraNet for the MS metric on this dataset.
Extensive experimental evaluations demonstrate that the MS metric constructed by the HydraNet framework provides actionable insights into how momentum impacts outcomes at different granularities, establishing a new foundation for momentum modeling and sports analysis. To the best of our knowledge, this is the first work to explore and effectively model momentum across multiple granularities in professional tennis tournaments. The source code and datasets are available at https://github.com/ReyJerry/HydraNet
.
\end{abstract}

\begin{keyword}
Sports Analytics \sep Momentum \sep Hydra \sep Multi-Granularity \sep Tennis Dataset
\end{keyword}





\end{frontmatter}


\section{introduction}
In recent years, sports competition analysis has become a central focus in sports science~\cite{jones2022research}. Traditional methods, which rely on empirical summarization, statistical analysis, and manual video interpretation~\cite{o2009research}, offer valuable insights but struggle to handle the increasing complexity of sports data. The advent of deep learning, with its ability to automatically extract meaningful patterns from high-dimensional and complex data, has provided new solutions for sports analysis~\cite{ghosh2023sports}, benefiting sports such as football~\cite{constantinou2019dolores, herold2019machine, sheng2020greensea, fernandez2021soccermap, robberechts2023xpass}, basketball~\cite{zhao2023using, fu2023hoopinsight, liu2025eitnet}, badminton~\cite{wang2023shuttleset, ibh2024stroke}, and tennis~\cite{ingram2019point, zhou2023using, giles2023differentiating}. These approaches have introduced new perspectives and methodologies for predicting match outcomes, analyzing tactics, and evaluating player performance. Among these applications, event-stream and tabular data, with their structured and event-driven nature, are particularly suited for small-ball sports analysis. These sports typically involve fewer participants, with outcomes more directly influenced by individual performance~\cite{cui2019data}. Additionally, their higher action frequencies and shorter rally durations result in highly detailed and granular match data~\cite{carboch2019match, lisi2024distribution}.

As a representative of small-ball sports, tennis matches are typically recorded in multiple granularities, including points, games, sets, and matches. Each rally can be described through specific events (e.g., serve, return, volley), closely linked to multidimensional factors such as a player's technical actions, physical condition, and psychological dynamics. The diversity and structured nature of tennis data provide ideal research conditions for match analysis~\cite{takahashi2022performance}. However, despite significant progress in predicting tennis match outcomes and analyzing player performance~\cite{sipko2015machine, makino2020feature, bayram2021predicting}, the critical phenomenon of momentum remains underexplored.

Momentum, defined as the dynamic performance trend of players during matches, is considered as a key factor influencing pivotal turning points and final outcomes~\cite{higham2000momentum, dietl2017momentum}. 
\begin{figure}[t]
    \centering
    \includegraphics[width=0.54\textwidth]{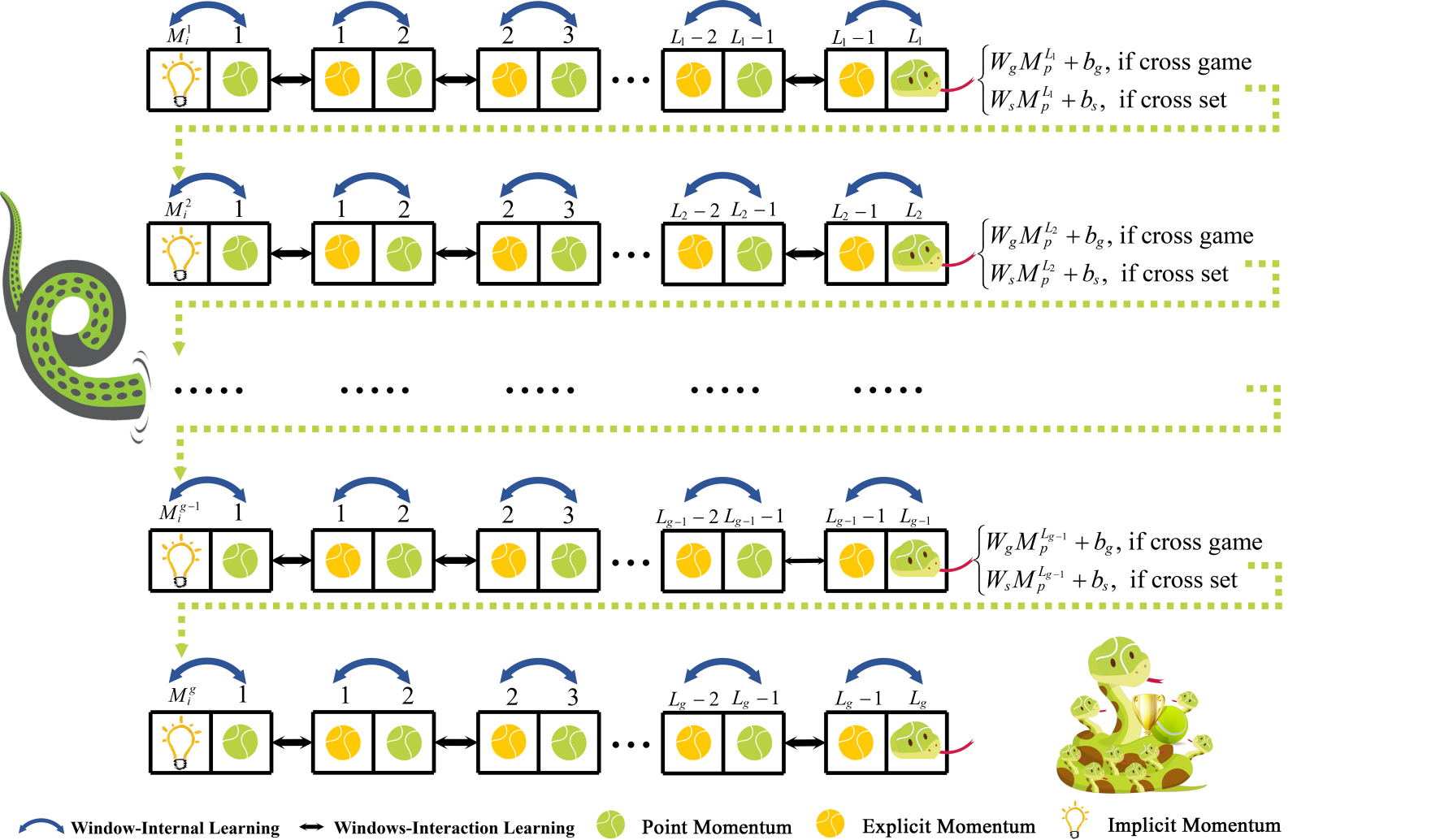}
    \caption{The conceptual diagram of \( M_s \) modeling using the Hydra methodology with \( M_p \), \( M_e \), and \( M_i \).}
    \label{fig:0}
\end{figure}

It adds unpredictability to sports competitions, keeping spectators engaged and making matches more exciting. However, modeling and analyzing momentum remain challenging due to its dynamic and multi-dimensional nature, which cannot be fully captured by static features or simplistic models. 
Although some studies have analyzed tennis matches with momentum~\cite{lei2024rhythms, wang2024multidimensional}, limitations still persist. First, they often utilize simplistic models with limited features, relying on basic weighted sums or decay curves for momentum construction, lacking precision and dynamic adaptability. Second, they fail to integrate established momentum theories in tennis, resulting in insufficient theoretical grounding. Third, the datasets are typically narrow, such as focusing on the first two rounds of the 2023 Wimbledon men's singles, which limits sample size and undermines generalizability. Additionally, most analyses focus on single-granularity evaluations, neglecting momentum's effects across different competition levels (games, sets, matches). These gaps underscore the need for more comprehensive, precise, and data-rich momentum modeling.

To address these challenges, we propose a MS metric for multi-granularity tennis tournaments analysis and a momentum-driven SSD-based framework, named HydraNet, designed for the modeling of MS. Additionally, we construct a million-level tennis cross-tournament dataset to validate HydraNet's capability in MS modeling. Specifically, this paper makes contributions in terms of evaluation metric, methodologies, and datasets for tennis momentum analysis across tournaments:
\begin{enumerate}
    \item We propose a novel metric, MS, for modeling the momentum phenomenon across multiple granularities in tennis tournaments;

    \item We propose HydraNet, a novel momentum-driven SSD framework, which comprises: (i) the Hydra module, designed to capture both explicit and implicit momentum through fine-grained sliding-window and cross-game state propagation mechanisms based on the SSD framework; (ii) the Versus Learning module, which models the adversarial nature of tennis matches at a macro level to enhance the complementarity and diversity of player momentum; (iii) the CAAM module, which captures and integrates complex momentum interactions at a micro level, both within and between players, aligning momentum modeling with real-world competitive dynamics.
    



    \item We construct a large-scale tennis cross-tournament dataset, comprising millions of data points from the 2012–2023 Wimbledon and 2013–2023 US Open, and use it to validate HydraNet's ability to model the MS metric across multiple granularities tennis tournaments;

    \item Through MS metrics in our experiments, we identified interesting trends, such as the 'half-time champagne' phenomenon, and the varying impact of multi-source information on matches at different granularities.

\end{enumerate}
\section{DATASET BUILDING}
To ensure the generalization and credibility of our experimental results, we curated data from the official Grand Slam websites, covering most matches from the Wimbledon Championships (2012–2023) and the US Open Tennis Championships (2013–2023). Following rigorous data cleaning and standardization, our dataset spans these two prestigious tournaments, consisting of 1,021,178 points, 162,051 games, 16,649 sets, and 5,712 matches played by 883 professional players, including 441 male players (e.g., Rafael Nadal, Novak Djokovic, Roger Federer) and 442 female players (e.g., Serena Williams, Iga Swiatek, Ashleigh Barty). Specifically, the Wimbledon Dataset (WID) contains 561,760 points, 89,646 games, 9,065 sets, and 3,069 matches, while the USOpen Dataset (USD) includes 459,418 points, 72,405 games, 7,584 sets, and 2,643 matches. All matches in the datasets adhere to the Grand Slam rules: best-of-five sets for men’s singles and best-of-three sets for women’s singles. Each set typically consists of six games or more, and a tiebreaker is played if the score reaches 6-6. Each point is described in detail using metadata that captures the fine-grained dynamics from serve to the conclusion of the rally, including 54 player-specific features that comprehensively capture player performance, match dynamics, and key indicators. A complete definition of these features, along with details of the dataset construction process, is provided in Appendix A.1.
\section{MEASURING MOMENTUM}
In this study, we propose a novel metric, denoted as MS, to model the momentum phenomenon across multiple granularities in tennis tournaments. The representation of the current Momentum Score MS is obtained through two steps: self-momentum modeling and adversarial relationship learning.

\textbf{Self-momentum modeling}. 
Based on the insights of Iso-Ahola et al.~\cite{iso2016psychological}, who proposed that overall performance consists of occurrences of momentum that vary in frequency and duration, we define three types of momentum that impact player performance to varying degrees during a match, and introduce the \( \mathrm{Hydra}(\cdot) \) algorithm to model the interactions among these momenta. The associated formula is as follows:
\begin{equation}
    M_s = \mathrm{Hydra}(M_p, M_e, M_i)
\end{equation}
where \( M_p \) denotes the point momentum, which reflects the momentum gained from the performance of the current point. \( M_e \) represents the explicit momentum, while \( M_i \) indicates the implicit momentum. \( M_s \) represents the self-momentum constructed through \( M_p \), \( M_e \) and \( M_i \). The illustrative computation process of \( M_s \) is shown in Figure 1. 

Building upon existing theoretical research in tennis~\cite{moss2015momentum,brody1987tennis,johnson2006performance,vernon2018returning,colomar2022determinant,harris2021psychological,hornery2007fatigue}, 
we define point momentum (\( M_p \)) as a multifaceted construct, comprising serve, return, psychology, and fatigue. Each component is represented by player performance features, as detailed in Table~\ref{tab:1}, which encompasses a total of 32 factors. 
Additionally, we recognize that point momentum is influenced not only by the player's current performance but also by the match's historical progression, which we refer to as historical momentum. Inspired by Iso-Ahola et al.~\cite{iso2016psychological}, who also propose that momentum evolves from a conscious to an unconscious driver of behavior, we argue that behavior-driven point momentum (\( M_p \)) could transform into different forms of historical momentum over varying time scales, comprising two types: \textit{explicit momentum} \( M_e \), stemming from the previous point's performance and outcome, and \textit{implicit momentum} \( M_i \), which comes into play when a player enters a new game or set. \( M_i \) is typically shaped by the final outcome of the previous game or set, as well as adjustments made by the player, including rest, strategy, and coaching guidance between games or sets.
\begin{table}[t]
    \caption{Feature Set for Modeling Point Momentum Factors of Player1 and Player2 in Tennis Matches.}
    \begin{adjustbox}{width=0.48\textwidth}
    \scriptsize
    \begin{tabular}{@{}lccccc@{}}
        \toprule
        \textbf{Factor}     & \textbf{Serve}                              & \textbf{Return}                              & \textbf{Psychology}                          & \textbf{Fatigue}                             \\ \midrule
        \textbf{Player1}  & \begin{tabular}[c]{@{}c@{}}p1\_serve \\ p1\_double\_fault \\ p1\_break\_pt\_missed \\ p1\_ace \\ p1\_serve\_speed \\ p1\_serve\_depth\end{tabular} 
                           & \begin{tabular}[c]{@{}c@{}}p1\_break\_pt\_won \\ p1\_return\_depth\end{tabular}        & \begin{tabular}[c]{@{}c@{}}p1\_unf\_err \\ p1\_net\_pt \\ p1\_net\_pt\_won \\ p1\_winner \\ p1\_points\_diff \\ p1\_game\_diff \\ p1\_set\_diff\end{tabular} 
                           & p1\_distance\_run                            \\ \midrule
        \textbf{Player2}  & \begin{tabular}[c]{@{}c@{}}p2\_serve \\ p2\_double\_fault \\ p2\_break\_pt\_missed \\ p2\_ace \\ p2\_serve\_speed \\ p2\_serve\_depth\end{tabular} 
                           & \begin{tabular}[c]{@{}c@{}}p2\_break\_pt\_won \\ p2\_return\_depth\end{tabular}        & \begin{tabular}[c]{@{}c@{}}p2\_unf\_err \\ p2\_net\_pt \\ p2\_net\_pt\_won \\ p2\_winner \\ p2\_points\_diff \\ p2\_game\_diff \\ p2\_set\_diff\end{tabular} 
                           & p2\_distance\_run                            \\ \bottomrule
    \end{tabular}
    \end{adjustbox}
    \label{tab:1}
\end{table}

\textbf{Adversarial relationship learning}. Building upon the nonlinear coupling between the performances of both players in a tennis tournament~\cite{palut2005dynamical}, we introduce the opponent's performance into the player's MS modeling process for the first time. The corresponding formula is as follows:
\begin{equation}
    MS = \mathrm{CAAM}(\mathrm{Versus Learning}(M_{s1}, M_{s2}))
\end{equation}
where \( M_{s1} \) and \( M_{s2} \) represent the self-momentum of the player and the opponent, respectively, learned by the Hydra module. The Versus Learning method enhances competitive information between players at a macro level, while the CAAM module captures and integrates interactions across multiple momentum dimensions, both intra- and inter-player, at a micro level, after enhancing the competitive information. 
This results in a more realistic momentum score that better aligns with competition dynamics.
\section{METHODOLOGY}
\subsection{Problem Formulation}
Let \( M \) denote the collection of all tennis matches, where each match \( m \in M \) consists of multiple sets \( S \), each set \( s \in S \) consists of multiple games \( G \), and each game \( g \in G \) consists of multiple points \( p \). For each player in match \( m \), we record a sequence of feature vectors \( \mathbf{X} = [\mathbf{X}_1, \mathbf{X}_2, \ldots, \mathbf{X}_N] \), where \( N \) represents the total number of points played throughout the entire match. Each feature vector \( \mathbf{X}_i \in \mathbb{R}^{d} \) represents the in-game features of a player, where \( d \) denotes the number of features. Each point \( p \), game \( g \), set \( s \), and match \( m \) is associated with a binary label \( y_p \in \{0, 1\} \), \( y_g \in \{0, 1\} \), \( y_s \in \{0, 1\} \), and \( y_m \in \{0, 1\} \), respectively. The label of \( y = 1 \) indicates a positive outcome (e.g., player1 wins the point, game, set, or match), while \( y = 0 \) indicates a opposite outcome. The goal of this study is to develop a model that can capture the MS at each moment during a tennis match and use this MS to predict the outcomes of subsequent points, games, sets, and the overall match, thereby uncovering the potential of momentum in multi-granularity tennis tournaments analysis.
\subsection{Model Architecture}
\begin{figure*}[t]
\centering
\includegraphics[width=\textwidth]{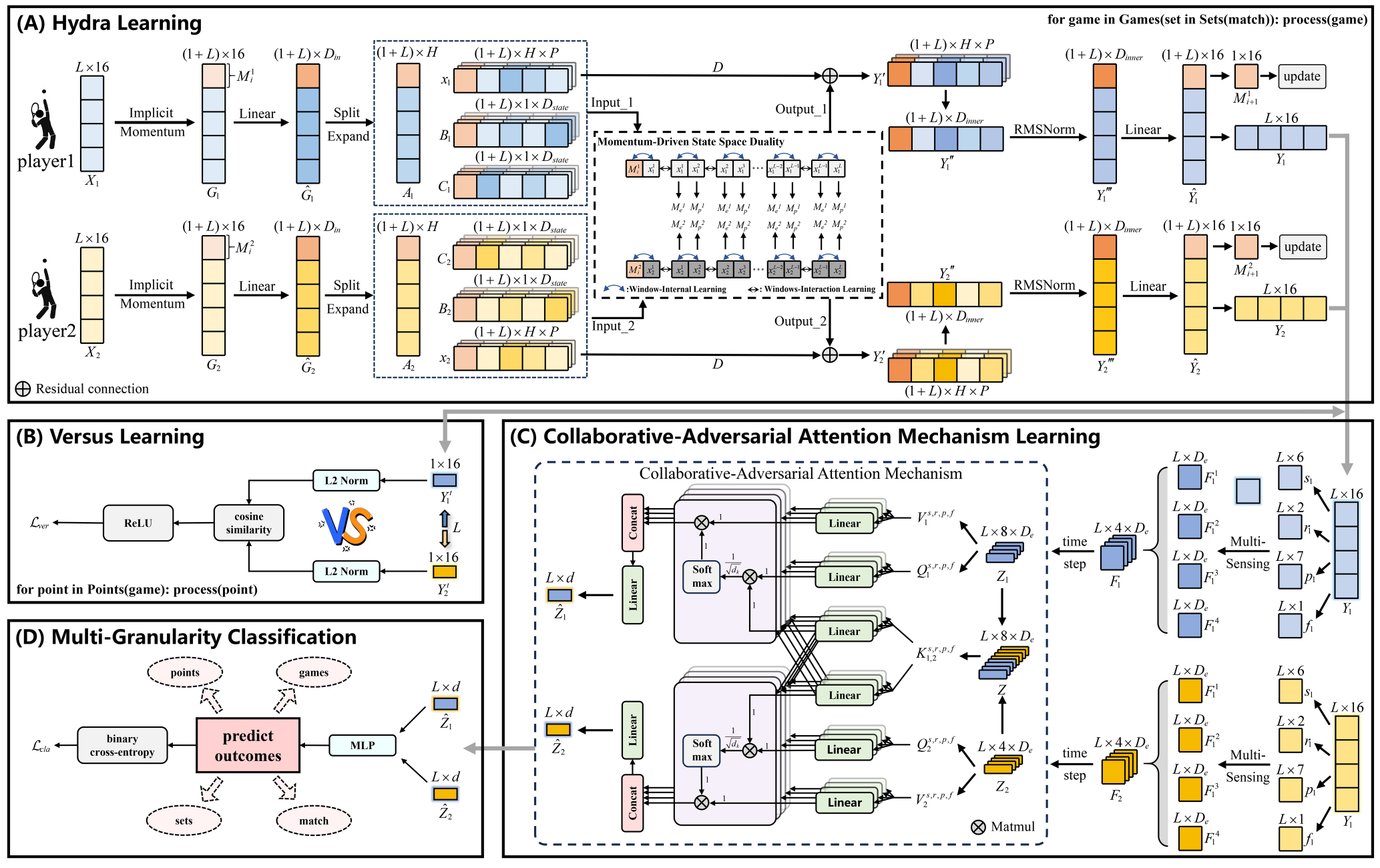}
\caption{Workflow of HydraNet: (a) Hydra Learning; (b) Versus Learning; (c) Collaborative-Adversarial Attention Mechanism learning; (d) Multi-Granularity Classification.               
                                            }
\label{fig:1}
\end{figure*}

HydraNet is a state space model comprising four main stages: (a) Hydra learning; (b) Versus Learning; (c) Collaborative-Adversarial Attention Mechanism learning; (d) Multi-Granularity Classification. In the following sections, we will review each stage in detail and provide the necessary operational specifics. The model architecture is illustrated in Figure~\ref{fig:1}.
\subsection{Hydra learning}
In the Hydra module, match data for player1 and player2 are first split by games and input sequentially in chronological order. The data is concatenated with their respective \( M_i \), followed by linear transformations and dimensional adjustments to generate dynamic features, temporal decay factors, input-to-state mappings, and state-to-output mappings. These components are processed within the Momentum-Driven State Space Duality (MSSD) framework, enabling both intra- and inter-window momentum learning. The final output captures \( M_e \) and \( M_i \), along with long- and short-term dependencies in the match sequence, reflecting the players' momentum.

\subsubsection{Implicit Momentum Module.} Considering the \textit{unconscious} nature of the long-range momentum~\cite{iso2016psychological}, we introduce an \textit{implicit momentum} module for both player1 and player2, denoted as \( M_i^1, M_i^2 \in \mathbb{R}^{1 \times d} \). These momentum representations are updated at the end of each game based on the final momentum data of player1 and player2. Additionally, the updates incorporate varying degrees of perturbation depending on whether transitions occur across games or sets:
\begin{equation}
\begin{cases}
M_i^1 = W_g^1 X_{\text{final}}^1 + b_g^1,  \quad M_i^2 = W_g^2 X_{\text{final}}^2 + b_g^2,  & \text{if cross-game}\\
M_i^1 = W_s^1 X_{\text{final}}^1 + b_s^1,  \quad M_i^2 = W_s^2 X_{\text{final}}^2 + b_s^2,  & \text{if cross-set}
\label{eq:2}
\end{cases}
\end{equation}
where \( W_g^1 \) and \( W_g^2 \) represent the learnable cross-game implicit momentum adjustment matrices for player1 and player2, respectively, and \( b_g^1 \) and \( b_g^2 \) are the corresponding learnable cross-game adjustment parameters. Similarly, \( W_s^1 \) and \( W_s^2 \) denote the learnable cross-set implicit momentum adjustment matrices, while \( b_s^1 \) and \( b_s^2 \) are the learnable cross-set implicit momentum adjustment parameters.

When the data for player1 and player2 is input, \( M_1^i \) and \( M_2^i \) are concatenated with the game data of player1 and player2, resulting in the game-level representations \( G_1 = [M_i^1, X_1^1, X_1^2, \dots, X_1^L] \in \mathbb{R}^{(1+L) \times d} \) and \( G_2 = [M_i^2, X_2^1, X_2^2, \dots, X_2^L] \in \mathbb{R}^{(1+L) \times d} \), respectively. Here, \( X_1^i \in \mathbb{R}^{1 \times d} \) and \( X_2^i \in \mathbb{R}^{1 \times d} \) denote the performance data of player1 and player2 for the \( i \)-th point within the game, while \( L \) represents the number of points in the game. Through the cross-game propagation of \( M_i \), the model achieves the long-range momentum effects modeling across games and even across sets.

\subsubsection{Construction of Core Parameters.}
To prepare for MSSD learning, we construct four critical components for each player: \( x \), \( \mathbf{A} \), \( \mathbf{B} \), and \( \mathbf{C} \). Specifically, \( x \) represents the core dynamic feature matrix, capturing the key attributes of the player’s performance. \( \mathbf{A} \) encodes the temporal decay factor, which models the natural degradation of momentum due to the time intervals between successive points. \( \mathbf{B} \) describes the input-to-state mapping matrix, projecting the raw features of each point in the tennis match into the latent state space, quantifying the immediate impact of various factors on a player's momentum. \( \mathbf{C} \) represents the state-to-output mapping matrix, translating the latent states into observable momentum indicators to support match prediction and tactical attribution. These components are essential for capturing the dynamic temporal information inherent in tennis matches and enabling efficient state-space transformations within the context of the match dynamics. The complete definition and computation of these key components is provided in Appendix A.2.

\subsubsection{Momentum-Driven State Space Duality.}
In this module, the primary objective is to integrate explicit and implicit momentum, along with recursive state updates across windows, through five steps to develop dynamic momentum modeling for sequential inputs. To avoid redundancy, all subsequent equations presented within the Hydra module are general formulas applicable to both players.

\textbf{The first step} involves segmenting the input sequence into localized temporal windows using a sliding window mechanism. By applying the \textit{unfold} operation, the input tensors \(\mathbf{x}, \mathbf{A}, \mathbf{B}, \mathbf{C}\) are partitioned into segments. The window size is defined as \( S_{\text{window}} = 2 \), and the total number of windows is \( N_{\text{window}} = L \). This segmentation ensures that the first window captures both the implicit momentum from the previous game and the point momentum from the first point of the current game, modeling the long-range effects of momentum, while each subsequent window captures the explicit momentum  \( M_e \) at the current time step, along with its corresponding point momentum \( M_p \), thereby ensuring that the explicit influence of momentum from the previous point is captured for each point in subsequent calculations. After segmentation, the resulting tensor \(\mathbf{x}_{\text{windows}} \in \mathbb{R}^{N_{\text{window}} \times S_{\text{window}} \times H \times P}\) is expressed for each independent window \( x_{\text{window}} \) as:
\begin{equation}
\mathbf{x}_{\text{window}} =
\begin{cases} 
\{M_i, x_1\}, & \text{if } n = 1 \\
\{x_{n-1} \to M_e, x_n \to M_p\}, & \text{if } n > 1
\end{cases}
\end{equation}
where , for \( n > 1 \), \( X_{n-1} \) represents the state of \( M_e \), while \( x_{n} \) represents the state of \( M_p \). During subsequent window-wise learning, each window updates the state of \( x_{n-1} \) for the following window. This mechanism facilitates the dynamic propagation of momentum across windows. Similarly, the \textit{unfold} operation is applied to \(\mathbf{A}, \mathbf{B}, \mathbf{C}\) to generate their respective local representations within the sliding window, expressed as \( \mathbf{A}_{\text{windows}} \in \mathbb{R}^{H \times N_{\text{window}} \times S_{\text{window}}}\), while \( \mathbf{B}_{\text{windows}}, \mathbf{C}_{\text{windows}} \in \mathbb{R}^{{N_{\text{window}}} \times S_{\text{window}} \times 1 \times D_{\text{state}}}\). These local representations serve as the foundation for capturing fine-grained temporal dynamics and enable subsequent computations to focus on localized interactions while maintaining global temporal consistency.

\textbf{The second step}  
focuses on modeling the nonlinear cumulative effects of between adjacent points momentum within the window through the state transition matrix \( \mathbf{L} \), capturing the macroscopic momentum interaction mechanisms (e.g., advantage at the end of the previous game \(\to\) boosting the player's determination for the first point of the current game, or score on point t-1 \(\to\) increased aggression on point t) between implicit momentum \(M_i\), explicit momentum \(M_e\), and point momentum \(M_p\). 
Building on the segmented windows from the sliding window mechanism, this step captures fine-grained momentum temporal dependencies while ensuring computational efficiency. Inspired by the attention mechanism, it emphasizes temporal interactions but uses a structured approach to reduce computational overhead and improve interpretability. Specifically, we compute \( \mathbf{L} \in \mathbb{R}^{H \times N_{\text{window}} \times S_{\text{window}} \times S_{\text{window}}} \), a matrix encoding cumulative temporal dynamics and attention head interactions across time steps, defined as:
\begin{equation}
L = \exp\left(S\right), \quad S_{ij} = 
\begin{cases} 
\sum_{k=i}^j \mathbf{A}_{\text{window}}^k, & \text{if } i \leq j \\
-\infty, & \text{if } i > j
\end{cases}
\end{equation}
where \( S \in \mathbb{R}^{H \times N_{\text{window}} \times S_{\text{window}} \times S_{\text{window}}} \) is the cumulative sum matrix constructed along the temporal dimension for each sliding window, and 
\( S_{ij} \) represents the cumulative sum from time step \( i \) to time step \( j \) within the corresponding sliding window. \(A_{\text{window}} \in A_{\text{windows}} \). Leveraging the matrix \( L \), the diagonal output \( Y_{\text{diag}} \) for each window is computed by aggregating momentum features and feature mappings within the sliding windows using the Einstein summation convention (einsum), as follows:
\begin{equation}
Y_{\text{diag}} =  \sum_{s=1}^{S_{\text{window}}}(\sum_{s=1}^{S_{\text{window}}}(\sum_{d=1}^{D_{\text{state}}}( \mathbf{C}_{\text{windows}} \cdot \mathbf{B}_{\text{windows}}) \cdot L) \cdot \mathbf{x}_{\text{windows}} \big)
\end{equation}
where \(n\) represents the index of the sliding window, while \(s\) denotes the time step index within each window. These temporal indices are involved in the accumulation and computation of interactions within the matrices \(\mathbf{C}_{\text{windows}}\), \(\mathbf{B}_{\text{windows}}\), and \(\mathbf{L}\), each of which captures the momentum relationships between the temporal steps within the sliding window.

\textbf{The third step} calculates the recursive intra-window states, quantifying the contribution weight of the historical state to the current window (e.g., positive feedback formed by consecutive serves won), preparing for subsequent state computations across windows. First, we compute the cumulative state \( A_{\text{cumsum}} \) by accumulating the time steps of \( A_{\text{windows}} \), as follows:
\begin{equation}
{A}_{\text{cumsum}} = \sum_{s=1}^{S_{\text{window}}} {A}_{\text{windows}}
\end{equation}
Next, the window's decayed state \( E \) is computed as follows:
\begin{equation}
E = \exp(\hat{A}_{\text{cumsum}} - {A}_{\text{cumsum}})
\end{equation}
where \( \hat{A}_{\text{cumsum}} \in \mathbb{R}^{H \times N_{\text{window}} \times 1}\) is the most recent time step of \( {A}_{\text{cumsum}} \in \mathbb{R}^{H \times N_{\text{window}} \times S_{\text{window}}}\).
The decayed state \( E \in \mathbb{R}^{H \times N_{\text{window}} \times S_{\text{window}}} \) is used to quantify the contribution weights of historical momentum to the current window. Based on this, the window state update equation is:
\begin{equation}
W_s = \sum_{s=1}^{S_{\text{window}}} ({B}_{\text{windows}} \cdot E \cdot {x}_{\text{windows}})
\end{equation}
where \( W_s \in \mathbb{R}^{N_{\text{window}} \times H \times P \times D_{\text{state}}} \) represents the window's state, updated by the decay weight, dynamically adjusting the point characteristics and generating momentum-dependent dynamic window patterns.

\textbf{The fourth step} 
focuses on capturing the microscopic momentum dependencies between non-adjacent points (e.g., break point advantage at point t-2 \( \to\) increased confidence in return at point t), by recursively propagating the long-term effects of momentum across windows.
First, we calculate the decay chunk \( F \) to prepare for the subsequent state updates across windows, as follows:
\begin{equation}
F = \exp\left(\hat{S}\right), \quad \hat{S_{ij}} = 
\begin{cases} 
\sum_{k=i}^j \mathbf{\tilde{A}}_{\text{cumsum}}^k, & \text{if } i \leq j \\
-\infty, & \text{if } i > j
\end{cases}
\end{equation}
where \( \tilde{A} \in \mathbb{R}^{H \times (N_{\text{window}} + 1)} \) is the result of first extracting the last time step from the \( {A}_{\text{cumsum}} \in \mathbb{R}^{H \times N_{\text{window}} \times S_{\text{window}}}\), and then padding the extracted values along the time dimension to introduce an additional entry at the beginning of the sequence. 
Then, we compute the SSM approximation for the spatial boundary during the state update process. We construct an initial windows' state \( W_{i} \in \mathbb{R}^{1 \times H \times P \times D_{\text{state}}}\) of all zeros, and concatenate it with the current windows' spatial state \( W_s \) to form the augmented windows' state \( W_{s'} \in \mathbb{R}^{N_{\text{window}} \times H \times P \times D_{\text{state}}}\). This concatenation ensures the global properties of the time-sequenced model are maintained. Then the spatial windows' state update is calculated as follows:
\begin{equation}
W_{s''} = \sum_{n=1}^{N_{\text{window} + 1}} (F \cdot W_{s'})
\end{equation}
where \( W_{s''} \in \mathbb{R}^{(N_{\text{window}} + 1) \times H \times P \times D_{\text{state}}} \) represents the updated windows' spatial state. Finally, the estimated windows' spatial state \( \tilde{W_s} \in \mathbb{R}^{N_{\text{window}} \times H \times P \times D_{\text{state}}} \) is obtained by selecting the relevant portion of the updated windows' state \( W_{s''} \). Then we combine the window-state \( \tilde{W_s} \) and window-time features, transforming the time-state dynamic information into the final momentum windows-interaction feature \( Y_{\text{off}} \), which includes both window-state and window-time dynamic momentum features. The formula is as follows:
\begin{equation}
Y_{\text{off}} = \sum_{d=1}^{D_{\text{state}}} {C_{\text{windows}}} \cdot \tilde{W_s} \cdot \exp({A}_{\text{cumsum}})
\end{equation}

\textbf{In the fifth step}, we use cross-attention mechanisms to capture the interaction between the within-window macroscopic momentum \( Y_{\text{diag}} \) and the across-windows microscopic momentum \( Y_{\text{off}} \). 
First, we restore \( {Y}_{\text{diag}} \) and \( {Y}_{\text{off}} \) from their windowed representations back to the original sequences \( \mathbb{R}^{(L + 1) \times H \times P} \). Next, the cross-attention mechanism is applied to fuse the window-internal and windows-interaction dynamic features, resulting in the fused feature \( Y \in \mathbb{R}^{(L + 1) \times H \times P} \):
\begin{equation}
Y = \text{softmax}\left( \frac{Q_{\text{diag}} K_{\text{off}}^T}{\sqrt{d_k}} \right) V_{\text{off}} + \text{softmax}\left( \frac{Q_{\text{off}} K_{\text{diag}}^T}{\sqrt{d_k}} \right) V_{\text{diag}}
\end{equation}
\subsubsection{Feature Fusion and Implicit Momentum Update}
First, we introduce the residual augmentation parameter \( D \) to balance the discrepancy between the current dynamic model output and the residual connection part, which is obtained by using State space duality (ssd) to get the output \( Y \) and the residual input \( \hat{x} \), enhancing information retention, thus yielding \( Y' = Y + x \cdot D \in \mathbb{R}^{(L + 1) \times H \times P} \). Next, the feature \( Y' \) is reconstructed as \( Y'' \in \mathbb{R}^{(L + 1) \times D_{dimer}} \), where the next step involves aggregating the features from multiple heads' dimensions \( H \) and each head's dimension \( P \). Subsequently, RMSNorm normalization is applied to \( Y'' \) to obtain \( Y''' \):
\begin{equation}
Y''' = \frac{Y'' \cdot (z \cdot \sigma(z))}{\sqrt{\sum_{j=1}^{D_{dinner}}(Y''_j)^2/D_{dinner} + \epsilon}} \cdot w
\end{equation}
where \( \sigma \) is the activation function, \( w \) is a learned normalization parameter, and \( \epsilon \) is a small constant value used to prevent instability. Then, \( Y'' \) is used for linear transformation to obtain the final output \( \hat{Y} \in \mathbb{R}^{(1 + L) \times d} \), which extracts the last time step information to update the implicit momentum \( M_i \).

Finally, we obtain the spatially coupled momentum modeling for player1 and player2, which integrates point momentum \( M_p \), explicit momentum \( M_e \), and implicit momentum \( M_i \). Subsequently, we utilize the latest values of the momentum for player1 and player2 at the final time step to update the implicit momentum \( M_i^1 \) and \( M_i^2 \) (as shown in Formula~\ref{eq:2}), facilitating subsequent cross-game or cross-set computations.
\subsection{Versus Learning}
In tennis, the inherent adversarial nature of the game ensures that one player prevails while the other is defeated, resulting in fundamentally contrasting behavioral patterns between the players. This contrast highlights the need to amplify the divergence in feature representations of adversarial entities. To formally capture this macroscopical phenomenon, we propose Versus Learning, a novel approach aimed at maximize the representation differences between two opposing entities, \( {y}_1^t \) and \( {y}_2^t \), which represent the momentum of player1 and player2, respectively, as output by the Hydra module at the \( t \)-th timestep.

HydraNet begins by normalizing the input features ${y}_1^t$ and ${y}_2^t$ using the L2 norm. We then measure their angular distance through cosine similarity and design a versus loss to enforce maximum separation:
\begin{equation}
\mathcal{L}_{\text{ver}} = \frac{1}{N} \sum \max \left( 0, m + \cos \left( \frac{{y}_1^t}{\|{y}_1^t\|_2}, \frac{{y}_2^t}{\|{y}_2^t\|_2} \right) \right)
\end{equation}
where \( m \) controls the separation margin. This approach fosters distinctly opposing behaviors for players, effectively capturing their contrasting momentum at a macro level, thereby providing a more complementary and diverse set of adversarial momentum features for subsequent feature fusion in the CAAM module.
\subsection{Collaborative-Adversarial Attention Mechanism Learning}
Given the interdependent and mutually influential nature of Serve, Return, Psychology, and Fatigue in a tennis match, which are affected by both a player’s performance and the opponent's feedback, we propose the CAAM. This mechanism captures and integrates the collaborative relationships among the four microcosmic factors within each player (e.g., the impact of physical condition on psychology) and the adversarial interactions between these microcosmic factors across both players (e.g., the effect of player2's Serve on player1's Return).
\subsubsection{Momentum Reconstruction Enhancement Module}
To enable targeted cross-dimensional learning in Collaborative-Adversarial Attention, we decompose the features output by the Hydra module along four microcosmic dimensions: serve index \( s \), return index \( r \), psychology index \( p \), and fatigue index \( f \). We first extract the features of player1 and player2 as specified in Table~\ref{tab:1}, obtaining ball-passing features \( s_1, s_2 \in \mathbb{R}^{L \times 6} \), reaction features \( r_1, r_2 \in \mathbb{R}^{L \times 2} \), mental features \( p_1, p_2 \in \mathbb{R}^{L \times 7} \), and fatigue features \( f_1, f_2 \in \mathbb{R}^{L \times 1} \). Then we use multi-sensing layers to enhance these feature representations, producing augmented embeddings \( F_1^1, \dots, F_1^4, F_2^1, \dots, F_2^4 \in \mathbb{R}^{L \times D_e} \), where \( D_e \) is the enhanced embedding dimension. The final augmented feature sets for player1 and player2 are then formed as \( F_1 = [F_1^1, \dots, F_1^4] \in \mathbb{R}^{L \times 4 \times D_e} \) and \( F_2 = [F_2^1, \dots, F_2^4] \in \mathbb{R}^{L \times 4 \times D_e} \).
\begin{table*}[htbp]
\centering
\footnotesize 
\caption{Performance Metrics of Multi-Granularity Momentum Prediction Results for Tennis Tournaments Based on HydraNet on the WID and USD.}
\begin{tabularx}{\textwidth}{*{8}{X}}
\toprule
Dataset & Granularity & AUC & AUPRC & ACC & F1-score & Recall & Precision \\
\midrule
\multirow{4}{*}{\centering WID} & Point & 0.9919$\pm$0.0024 & 0.9922$\pm$0.0024 & 0.9504$\pm$0.0099 & 0.9500$\pm$0.0103 & 0.9410$\pm$0.0158 & 0.9593$\pm$0.0099 \\
& Game & 0.8130$\pm$0.0046 & 0.7916$\pm$0.0019 & 0.7803$\pm$0.0061 & 0.7808$\pm$0.0049 & 0.7791$\pm$0.0090 & 0.7825$\pm$0.0025 \\
& Set & 0.6749$\pm$0.0108 & 0.6395$\pm$0.0104 & 0.6362$\pm$0.0077 & 0.6272$\pm$0.0177 & 0.6243$\pm$0.0141 & 0.6313$\pm$0.0078 \\
& Match & 0.9511$\pm$0.0054 & 0.9526$\pm$0.0070 & 0.8768$\pm$0.0128 & 0.8710$\pm$0.0192 & 0.8882$\pm$0.0133 & 0.8559$\pm$0.0115 \\
\midrule
\multirow{4}{*}{\centering USD} & Point & 0.9964$\pm$0.0005 & 0.9965$\pm$0.0005 & 0.9658$\pm$0.0034 & 0.9662$\pm$0.0033 & 0.9755$\pm$0.0036 & 0.9571$\pm$0.0067 \\
& Game & 0.7780$\pm$0.0089 & 0.7514$\pm$0.0109 & 0.7392$\pm$0.0057 & 0.7478$\pm$0.0032 & 0.7551$\pm$0.0110 & 0.7407$\pm$0.0060 \\
& Set & 0.6654$\pm$0.0127 & 0.6767$\pm$0.0115 & 0.6519$\pm$0.0130 & 0.6538$\pm$0.0182 & 0.6461$\pm$0.0105 & 0.6617$\pm$0.0138 \\
& Match & 0.9465$\pm$0.0077 & 0.9524$\pm$0.0027 & 0.8685$\pm$0.0176 & 0.8781$\pm$0.0170 & 0.8796$\pm$0.0108 & 0.8770$\pm$0.0122 \\
\bottomrule
\end{tabularx}
\label{tab:performance_metrics}
\end{table*}
\subsubsection{Collaborative-Adversarial Attention Mechanism}
At each time step \( t \), we obtain the microcosmic feature tensors \( Z_{t1} \) and \( Z_{t2} \) for player1 and player2, respectively. We then stack these features as \( Z_t \) and use the four microcosmic feature dimensions of both players as Queries to query their own and the opponent's microcosmic feature groups as Keys. The query information is used to update the player’s vector, capturing both the collaborative relationships within a player and the adversarial relationships between players. The features are aggregated based on the attention weights to obtain \( \hat{Z}_t \). The specific formula is as follows:
\begin{equation}
\hat{Z}_i = \sum_{m \in \{S, R, P, F\}} \text{softmax} \left( \frac{Q_{i,m} \left( K_{i,m} \cup K_{j,m} \right)^T}{\sqrt{d_k}} \right) V_{i,m}
\end{equation}
where \( i, j \in \{1, 2\}, i \neq j \) ensures that the interactions between the two players' behaviors are learned. \( \sum_{m \in \{S, R, P, F\}} \) indicates the inclusion of microcosmic features such as serve, receive, psychology, and fatigue in the aggregation. \( Q_i \) represents the player’s query and \( K_{i,m} \) and \( K_{j,m} \) represent the keys of player1 and player2 respectively. These keys are used for querying and aggregating microcosmic information from the collaborative and adversarial relations between the players. Finally, the players’ integrated attention results are computed as \( \hat{Z}_1, \hat{Z}_2 \in \mathbb{R}^{1 \times d} \).
\subsection{Multi-Granularity Classification}
To validate whether momentum has a multi-granularity impact on tennis matches, we attempt to make multi-granularity predictions using momentum across points, games, sets, and matches.

First, the final features \( \hat{Z_1} \) and \( \hat{Z_2} \) of player1 and player2 are concatenated and passed through an MLP layer to obtain the predicted momentum value \( \hat{y_t} \) for the current point. Then, the computed final momentum value is used to perform multi-granularity predictions for points, games, sets, and matches. For points, we predict the outcome of each point based on the momentum of the current rally, assessing whether the momentum shown by a player can determine the winner of that specific point. For games, we use the momentum from the last point of each game to predict the next game's outcome. For sets, we predict the outcome of the next set based on the momentum from the last point of the current set. For matches, we employ a ``half-time champagne'' strategy, using the momentum from the last point in the first half to predict the match outcome. We then use the binary cross-entropy loss function to calculate the prediction loss. The formula is as follows:
\begin{equation}
\mathcal{L}_{cla} = - \frac{1}{N} \sum \left( y_{cla} \cdot \log(\sigma(\hat{y_t})) + (1 - y_{cla}) \cdot \log(1 - \sigma(\hat{y_t})) \right)
\end{equation}
where \( y_{cla} \) represents the true label corresponding to the predicted value \( \hat{y} \), and \( cla \in \{point, game, set, match\}\). Finally, we sum \( \mathcal{L}_{ver} \) and \( \mathcal{L}_{cla} \) to obtain the total loss \( \mathcal{L} \). The formula is as follows:
\begin{equation}
\mathcal{L} = \mathcal{L}_{ver} + \mathcal{L}_{cla}
\end{equation}
\section{Experiment}
In this study, we use the WID and USD datasets constructed by ourselves, which are randomly split by match units into training and test sets (20\% for testing). We employ five-fold cross-validation on the training sets, selecting high-performance models for final independent testing to ensure robust generalizability. Without deliberate parameter tuning, the Hydra layer is set to 1, the CAAM module employs 8 heads. and the learning rate is fixed at 0.001. To mitigate overfitting, a dropout mechanism with a probability of 0.1 is implemented. Model performance is assessed using AUC, AUPRC, accuracy, F1 score, recall, and precision. Experiments are repeated to ensure the robustness and reliability of the results.
\begin{figure*}[t]
\centering
\includegraphics[width=\textwidth]{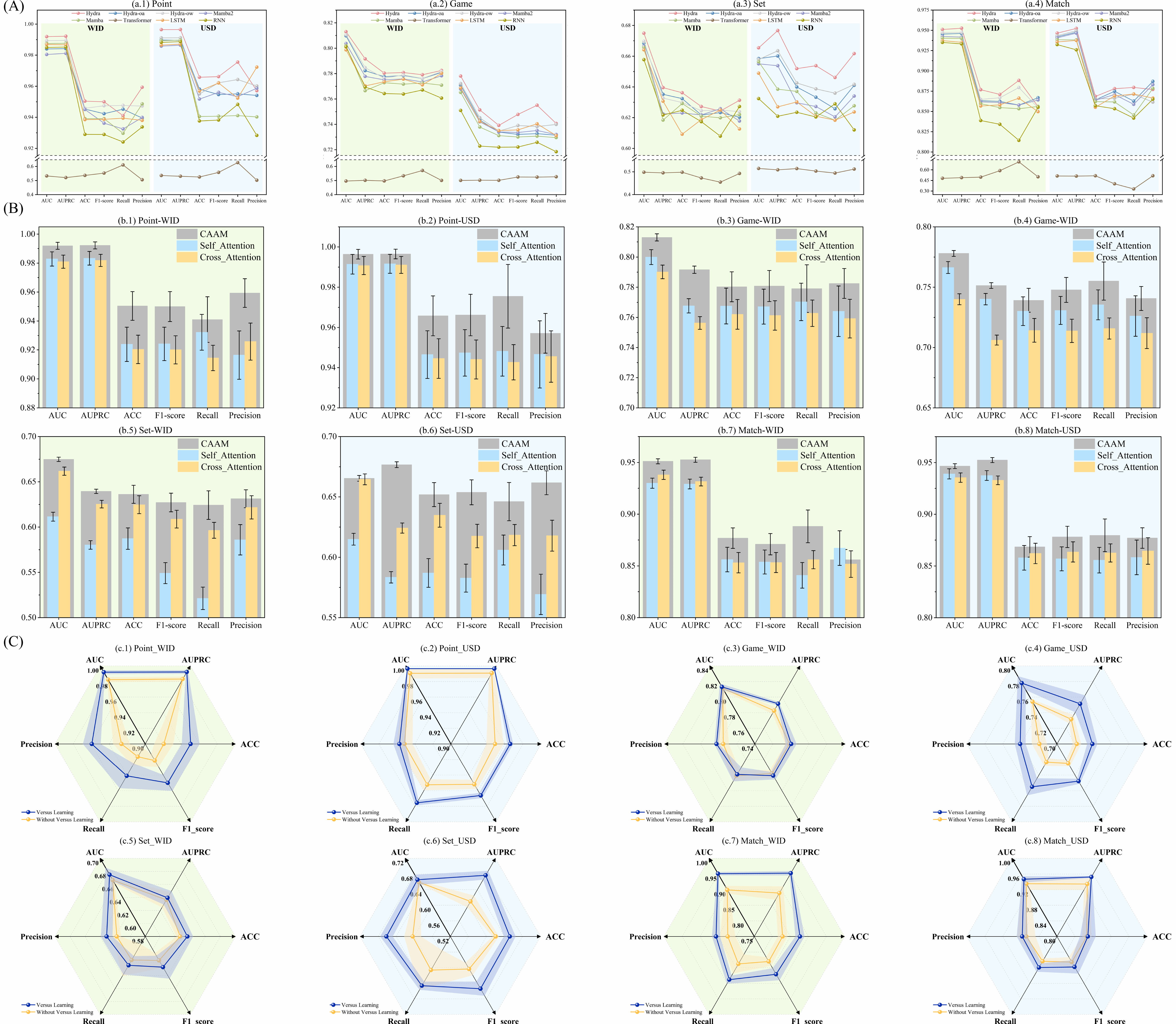}
\caption{Multi-Granularity Ablation Results for (a) Hydra module, (b) CAAM module and (c) Versus Learning method on the WID and USD Datasets.}
\label{fig:ablation}
\end{figure*}
\subsection{Multi-Granularity Performance Test}
The results of the multi-granularity performance comparative experiments are shown in Table~\ref{tab:performance_metrics}. We find that the MS metric constructed by HydraNet effectively predicts outcomes at the point, game, and match granularities, though its performance in set outcome prediction is relatively weaker. As outlined in subsection 4.6, HydraNet predicts the current point outcome based on its MS, the next game outcome using the MS of the previous game's last point, the next set using the MS of the last point of the previous set, and the match outcome using the ``half-time champagne'' strategy, predicting the result based on the MS of the last point from the first half of the match. Thus, HydraNet's MS indicator modeling at the point, game, and set granularities aligns with the expected modeling difficulty and achieves high performance. The precise prediction of match outcomes under the ``half-time champagne'' strategy is a surprising finding, as its span is longer than that of the game and set, yet it still reaches MS prediction strength comparable to the point granularity. Although direct evidence linking halftime performance to match outcomes in tennis is lacking, this study suggests that the last point of the first half may serve as a pivotal moment with significant influence on the match's trajectory and final result. If coaches and players leverage momentum management during the break to control the match's pace and psychological pressure, they can influence their own emotions and performance while disrupting the opponent's mental state, potentially gaining a momentum advantage that could determine the match outcome.


\subsection{Ablation Study}
For the HydraNet architecture, we design three sets of ablation experiments: Hydra ablation, CAAM ablation, and Versus Learning ablation, to thoroughly analyze HydraNet's MS modeling capabilities.
\begin{table*}[t]
\centering
\footnotesize 
\caption{Multi-Granularity Experimental Results for Multimodal Learning with Missing Modality Based on HydraNet on the WID.}
\begin{tabularx}{\textwidth}{*{9}{X}}
\toprule
Granularity & MLMM & AUC & AUPRC & ACC & F1$-$score & Recall & Precision & $P_{\text{combined}}$\\
\midrule
\multirow{4}{*}{\centering Point} 
& Serve & 0.9644$\pm$0.0093 & 0.9663$\pm$0.0085 & 0.8814$\pm$0.0194 & 0.8827$\pm$0.0176 & 0.8901$\pm$0.0151 & 0.8786$\pm$0.0226 & 2.390$e-$10\\
& Return & 0.9606$\pm$0.0049 & 0.9628$\pm$0.0046 & 0.8796$\pm$0.0070 & 0.8789$\pm$0.0068 & 0.8750$\pm$0.0110 & 0.8832$\pm$0.0109 & 7.092$e-$35\\
& Psychology & 0.8534$\pm$0.0183 & 0.8635$\pm$0.0149 & 0.7593$\pm$0.0207 & 0.7591$\pm$0.0221 & 0.7565$\pm$0.0191 & 0.7621$\pm$0.0281 & 0.000$e+$00\\
& Fatigue & 0.9855$\pm$0.0057 & 0.9860$\pm$0.0056 & 0.9310$\pm$0.0142 & 0.9309$\pm$0.0141 & 0.9291$\pm$0.0173 & 0.9329$\pm$0.0191 & 2.987$e-$01\\
\midrule
\multirow{4}{*}{\centering Game} 
& Serve & 0.7844$\pm$0.0134 & 0.7581$\pm$0.0133 & 0.7582$\pm$0.0202 & 0.7592$\pm$0.0200 & 0.7568$\pm$0.0196 & 0.7615$\pm$0.0204 & 1.652$e-$02\\
& Return & 0.7908$\pm$0.0108 & 0.7715$\pm$0.0121 & 0.7711$\pm$0.0096 & 0.7630$\pm$0.0088 & 0.7639$\pm$0.0066 & 0.7621$\pm$0.0111 & 1.121$e-$02\\
& Psychology & 0.7631$\pm$0.0125 & 0.7611$\pm$0.0128 & 0.7660$\pm$0.0131 & 0.7769$\pm$0.0156 & 0.7555$\pm$0.0123 & 0.0303$\pm$0.0245 & 6.703$e-$10\\
& Fatigue & 0.7927$\pm$0.0140 & 0.7668$\pm$0.0159 & 0.7632$\pm$0.0150 & 0.7645$\pm$0.0151 & 0.7635$\pm$0.0157 & 0.7655$\pm$0.0146 & 1.394$e-$01\\
\midrule
\multirow{4}{*}{\centering Set} 
& Serve & 0.6544$\pm$0.0129 & 0.6283$\pm$0.0131 & 0.6239$\pm$0.0208 & 0.6325$\pm$0.0154 & 0.6470$\pm$0.0195 & 0.6187$\pm$0.0135 & 6.400$e-$01\\
& Return & 0.6615$\pm$0.0151 & 0.6318$\pm$0.0133 & 0.6262$\pm$0.0106 & 0.6338$\pm$0.0136 & 0.6539$\pm$0.0114 & 0.6151$\pm$0.0186 & 5.459$e-$01\\
& Psychology & 0.5974$\pm$0.0152 & 0.5778$\pm$0.0176 & 0.5749$\pm$0.0168 & 0.5640$\pm$0.0121 & 0.5666$\pm$0.0119 & 0.5663$\pm$0.0127 & 1.140$e-$14\\
& Fatigue & 0.6562$\pm$0.0115 & 0.6205$\pm$0.0103 & 0.6277$\pm$0.0194 & 0.6315$\pm$0.0108 & 0.6444$\pm$0.0140 & 0.6190$\pm$0.0182 & 5.402$e-$01\\
\midrule
\multirow{4}{*}{\centering Match} 
& Serve & 0.9295$\pm$0.0105 & 0.9205$\pm$0.0164 & 0.8590$\pm$0.0145 & 0.8618$\pm$0.0159 & 0.8726$\pm$0.0318 & 0.8524$\pm$0.0217 & 2.700$e-$01\\
& Return & 0.9324$\pm$0.0194 & 0.9278$\pm$0.0280 & 0.8627$\pm$0.0160 & 0.8662$\pm$0.0173 & 0.8810$\pm$0.0352 & 0.8531$\pm$0.0228 & 9.048$e-$01\\
& Psychology & 0.7470$\pm$0.0155 & 0.7201$\pm$0.0176 & 0.6943$\pm$0.0167 & 0.7019$\pm$0.0178 & 0.7268$\pm$0.0162 & 0.7130$\pm$0.0188 & 0.000$e+$00\\
& Fatigue & 0.9165$\pm$0.0098 & 0.9195$\pm$0.0096 & 0.8409$\pm$0.0090 & 0.8201$\pm$0.0174 & 0.8089$\pm$0.0187 & 0.8513$\pm$0.0097 & 5.879$e-$07\\
\bottomrule
\end{tabularx}
\label{tab:3.5}
\end{table*}
\begin{table*}[t!]
\centering
\footnotesize 
\caption{Multi-Granularity Experimental Results for Multimodal Learning with Missing Modality  Based on HydraNet on the USD.}
\begin{tabularx}{\textwidth}{*{9}{X}}
\toprule
Granularity & MLMM & AUC & AUPRC & ACC & F1-score & Recall & Precision & $P_{\text{combined}}$\\
\midrule
\multirow{4}{*}{\centering Point} 
& Serve & 0.9701$\pm$0.0135 & 0.9712$\pm$0.0126 & 0.9181$\pm$0.0284 & 0.9185$\pm$0.0283 & 0.9208$\pm$0.0286 & 0.9162$\pm$0.0281 & 1.989$e-$03\\
& Return & 0.9791$\pm$0.0083 & 0.9795$\pm$0.0078 & 0.9419$\pm$0.0205 & 0.9421$\pm$0.0205 & 0.9436$\pm$0.0247 & 0.9409$\pm$0.0239 & 2.220$e-$02\\
& Psychology & 0.8894$\pm$0.0124 & 0.8958$\pm$0.0118 & 0.7869$\pm$0.0099 & 0.7864$\pm$0.0120 & 0.7866$\pm$0.0364 & 0.7883$\pm$0.0240 & 0.000$e+$00\\
& Fatigue & 0.9869$\pm$0.0080 & 0.9875$\pm$0.0075 & 0.9455$\pm$0.0178 & 0.9443$\pm$0.0179 & 0.9432$\pm$0.0223 & 0.9457$\pm$0.0189 & 1.769$e-$01\\
\midrule
\multirow{4}{*}{\centering Game} 
& Serve & 0.7546$\pm$0.0111 & 0.7258$\pm$0.0108 & 0.7257$\pm$0.0133 & 0.7254$\pm$0.0154 & 0.7214$\pm$0.0135 & 0.7295$\pm$0.0184 & 3.573$e-$02\\
& Return & 0.7629$\pm$0.0089 & 0.7401$\pm$0.0099 & 0.7291$\pm$0.0076 & 0.7305$\pm$0.0079 & 0.7316$\pm$0.0092 & 0.7295$\pm$0.0078 & 5.510$e-$02\\
& Psychology & 0.7399$\pm$0.0107 & 0.7069$\pm$0.0099 & 0.7266$\pm$0.0158 & 0.7246$\pm$0.0107 & 0.7228$\pm$0.0126 & 0.7266$\pm$0.0056 & 4.820$e-$05\\
& Fatigue & 0.7618$\pm$0.0106 & 0.7373$\pm$0.0070 & 0.7280$\pm$0.0142 & 0.7295$\pm$0.0145 & 0.7321$\pm$0.0188 & 0.7270$\pm$0.0167 & 2.941$e-$01\\
\midrule
\multirow{4}{*}{\centering Set} 
& Serve & 0.6551$\pm$0.0131 & 0.6617$\pm$0.0133 & 0.6303$\pm$0.0192 & 0.6253$\pm$0.0153 & 0.6130$\pm$0.0160 & 0.6407$\pm$0.0184 & 2.413$e-$01\\
& Return & 0.6546$\pm$0.0127 & 0.6572$\pm$0.0172 & 0.6271$\pm$0.0181 & 0.6280$\pm$0.0170 & 0.6231$\pm$0.0169 & 0.6345$\pm$0.0105 & 2.167$e-$01\\
& Psychology & 0.6112$\pm$0.0149 & 0.5863$\pm$0.0129 & 0.5929$\pm$0.0198 & 0.5776$\pm$0.0141 & 0.5744$\pm$0.0153 & 0.5911$\pm$0.0108 & 2.291$e-$16\\
& Fatigue & 0.6565$\pm$0.0117 & 0.6639$\pm$0.0141 & 0.6263$\pm$0.0152 & 0.6253$\pm$0.0092 & 0.6198$\pm$0.0108 & 0.6308$\pm$0.0074 & 5.996$e-$02\\
\midrule
\multirow{4}{*}{\centering Match} 
& Serve & 0.9193$\pm$0.0148 & 0.9124$\pm$0.0181 & 0.8606$\pm$0.0168 & 0.8558$\pm$0.0163 & 0.8413$\pm$0.0162 & 0.8711$\pm$0.0205 & 5.309$e-$02\\
& Return & 0.9210$\pm$0.0169 & 0.9152$\pm$0.0170 & 0.8653$\pm$0.0192 & 0.8603$\pm$0.0200 & 0.8444$\pm$0.0246 & 0.8772$\pm$0.0207 & 2.178$e-$01\\
& Psychology & 0.7563$\pm$0.0136 & 0.7130$\pm$0.0145 & 0.7227$\pm$0.0151 & 0.7238$\pm$0.0147 & 0.7470$\pm$0.0186 & 0.7117$\pm$0.0160 & 0.000$e+$00\\
& Fatigue & 0.9072$\pm$0.0164 & 0.8943$\pm$0.0178 & 0.8344$\pm$0.0122 & 0.8342$\pm$0.0162 & 0.8344$\pm$0.0170 & 0.8433$\pm$0.0122 & 1.725$e-$05\\
\bottomrule
\end{tabularx}
\label{tab:4}
\end{table*}
\subsubsection{Hydra Ablation}
We perform ablation replacements of the Hydra module in HydraNet using Hydra-only attention (Hydra-oa), Hydra-only window (Hydra-ow), Mamba2~\cite{dao2024transformers}, Mamba~\cite{gu2023mamba}, Transformer~\cite{vaswani2017attention}, Long Short-Term Memory (LSTM)~\cite{hochreiter1997long}, and Recurrent Neural Network (RNN)~\cite{zaremba2014recurrent}. The experimental results are shown in Figure~\ref{fig:ablation} (A). It can be observed that on both the WID and USD datasets, the Hydra module consistently achieves the best performance across all six metrics at the four granularities. This demonstrates the state-of-the-art capability of the Hydra module in self-momentum modeling. This is due to Hydra's use of fine-grained sliding windows for segmentation of the tournaments process, addressing the phenomenon in tennis matches where 'the current point is significantly influenced by the previous point.' By doing so, Hydra effectively captures the impact of the previous point on the current one. Additionally, a cross-attention mechanism dynamically captures both intra- and inter-block information, balancing short-term and long-term information influences throughout the match.

\subsubsection{CAAM Ablation}
We conduct an ablation analysis of the CAAM module using both cross-attention and self-attention modules. Experimental results are illustrated in Figure~\ref{fig:ablation} (B). It is clear that CAAM consistently outperforms the other two attention modules across all granularity levels and six metrics in both datasets. This is attributed to CAAM's consideration of the interdependencies at a micro level between a player's own Serve, Return, Psychology, and Fatigue, as well as the mutual influences of multi-source information between players. As a result, CAAM more accurately models the interaction of multi-source information both within and between players at a micro level during the match.
\subsubsection{Versus Learning Ablation}
We perform an ablation comparison between HydraNet with and without the Versus Learning method, as shown in Figure~\ref{fig:ablation} (C). Experimental results from both datasets reveal that HydraNet enhanced with Versus Learning demonstrates significantly stronger performance in multi-granularity momentum prediction tasks. This highlights the necessity of enhancing the adversarial relationships at a macro level between players during the MS construction process.






\subsection{Multimodal Learning with Missing Modality}
\begin{figure*}[t!]
\begin{minipage}[t]{0.5\textwidth}
\centering
\includegraphics[width=3.61in]{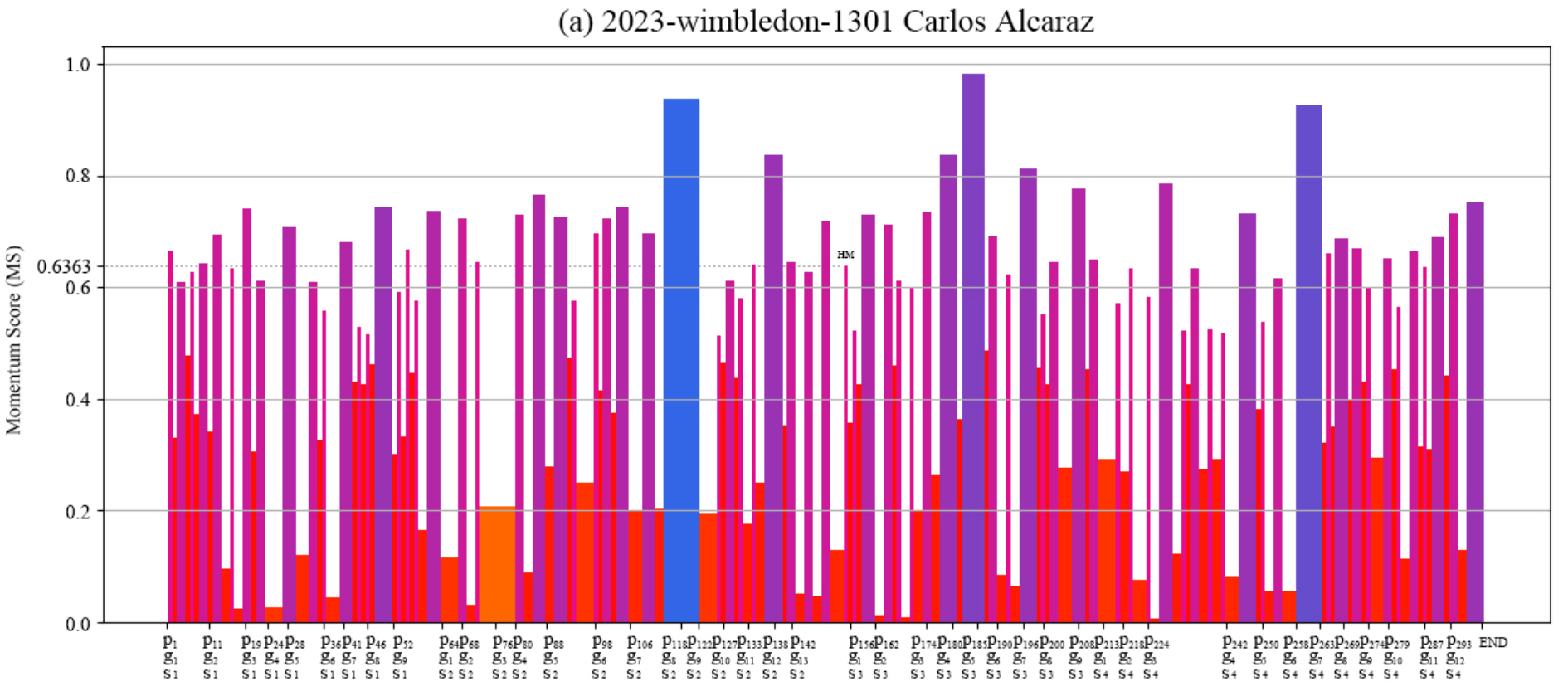}
\end{minipage}%
\begin{minipage}[t]{0.5\textwidth}
\centering
\includegraphics[width=3.61in]{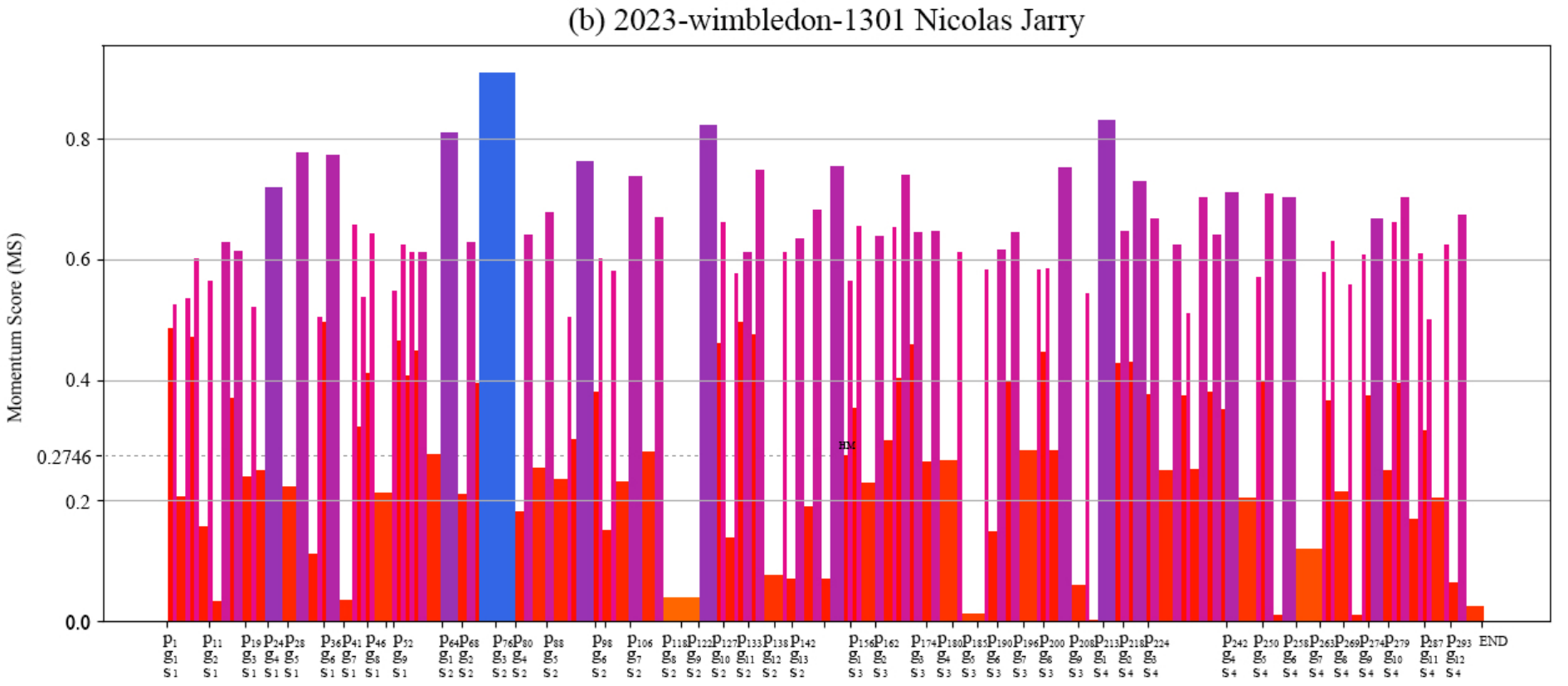}
\end{minipage}
\begin{minipage}[t]{0.5\textwidth}
\centering
\includegraphics[width=3.61in]{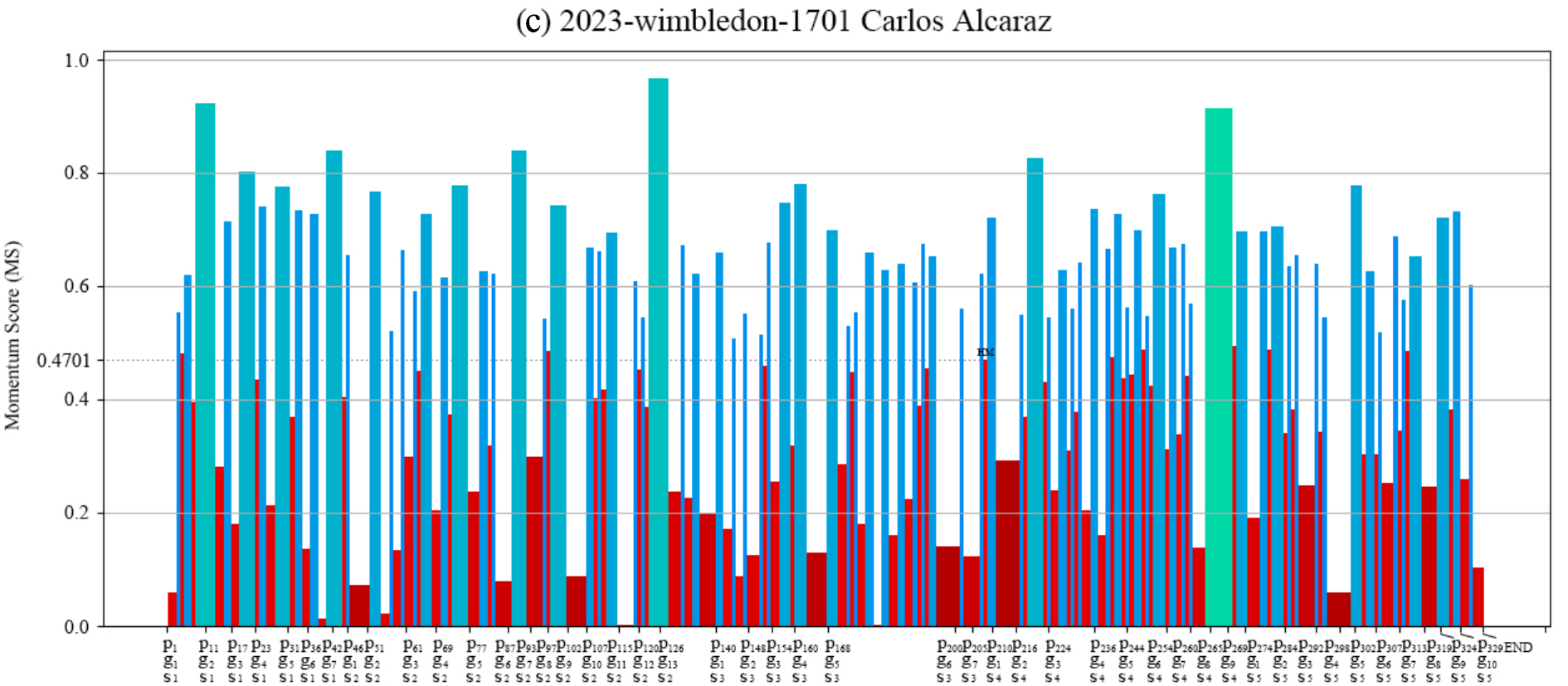}
\end{minipage}%
\begin{minipage}[t]{0.5\textwidth}
\centering
\includegraphics[width=3.61in]{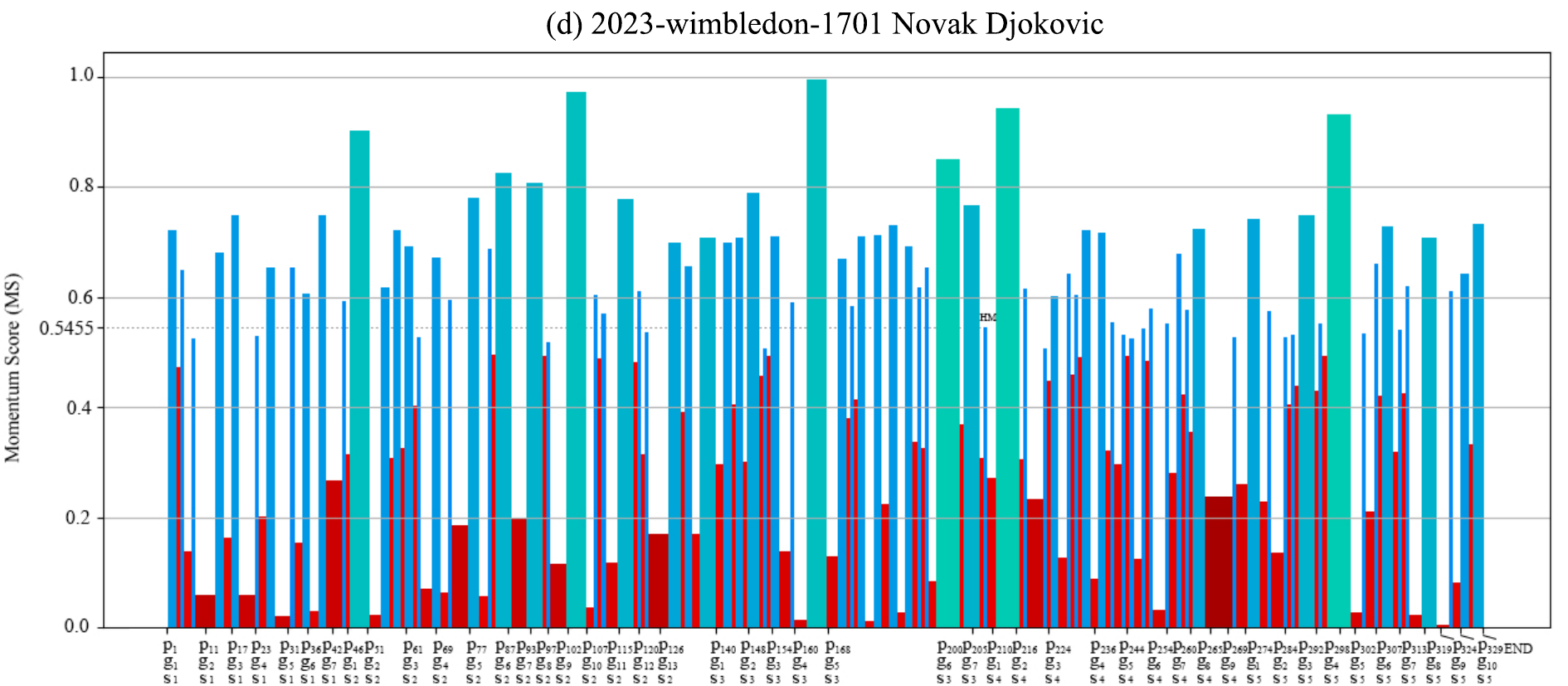}
\end{minipage}
\caption{A case study on the multi-granularity analysis capability of the MS metric constructed by HydraNet.}
\label{fig:6}
\end{figure*}
In order to validate the impact of different modalities on the multi-granularity MS modeling in tennis tournaments, we conducted Multimodal Learning with Missing Modality (MLMM) for Serve, Return, Psychology, and Fatigue across various granularities in WID and USD. To better evaluate the significance of each modality's impact on multi-granularity MS modeling, we calculated the p-values for six metrics before and after ablation for each group. We then used the Fisher method to combine the p-values of the six metrics, with the calculation formula given as follows:
\begin{equation}
P_{\text{combined}} = -2 \sum_{i=1}^{k} \ln(p_i)
\end{equation}
where \( p_i \) represents the p-value for each individual metric, and \( k \) is the number of metrics being combined. The result is the combined p-value, which is used to assess the overall statistical significance. Detailed experimental results are presented in Tables~\ref{tab:3.5} and~\ref{tab:4}. According to the principle ``when the p-value is less than 0.05, the result is typically considered statistically significant,'' we observe the following:
\begin{itemize}
    \item \textbf{Point Granularity}: At the point level, Serve, Return, and Psychology significantly affect the multi-granularity MS modeling. This indicates that for each individual point (such as the result of a single ball), the player's serving technique, returning ability, and psychological state play crucial roles in determining the course and outcome of the match. In high-pressure situations, the psychological state of a player can cause significant fluctuations in performance for each point, making psychology particularly important at this granularity.

    \item \textbf{Game Granularity}: At the game level, Serve, Return, and Psychology continue to show significant effects, which aligns with the results at the point level. This suggests that, in a complete game (or set), the player's serving and returning skills, as well as their psychological state, are key factors in determining the outcome. A player may influence the result of an entire game through strong serves, stable returns, and a positive psychological state.

    \item \textbf{Set Granularity}: At the set level, only the Psychology modality has a significant impact on MS modeling. This indicates that, across multiple games in a set, the player's psychological state becomes more decisive. Especially during long matches, maintaining psychological stability, avoiding anxiety and stress, may be critical for the overall outcome. The impact of serving and returning skills may diminish at this level, while psychological factors start to influence the performance within the set.
    
    \item \textbf{Match Granularity}: At the match level, both Psychology and Fatigue modalities significantly influence MS modeling. This shows that psychological factors and physical fatigue play a determining role in the outcome of an entire match. As the match progresses, the changes in the player's physical and psychological state can become the key factors that determine victory or defeat. Long matches exacerbate fatigue, while psychological factors govern how players perform and cope under fatigue, making these two modalities particularly important at this granularity.
\end{itemize}
In summary, the experimental results show that as the granularity level increases, the impact of psychological factors and fatigue on multi-granularity MS modeling becomes more significant, while the influence of technical modalities such as serving and returning decreases. This suggests that in higher-level matches, a player's psychological resilience and physical recovery capabilities often become more important than technical skills.
\subsection{Momentum Score Case Study}
To validate the multi-granularity momentum measurement capability of the MS metric and their predictive ability for consecutive wins and losses in real-world matches, we conduct MS visualization analyses for the 2023 Wimbledon matches 1301 (w-1301) and 1701 (w-1701) featuring Carlos Alcaraz (one of the most sportsmanship-driven players on tour) and his opponents, as shown in Figure~\ref{fig:6}. We average and fuse the MS values corresponding to consecutive wins and losses in the match intervals, allowing the MS bars to vividly depict these streaks. Additionally, we annotate all cross-game and cross-set points for comparison with the actual match progression, and we also label the half-match (HM) points, recording their MS values to analyze the interesting ``half-time champagne'' phenomenon.
First, we compare the consecutive win/loss segments in the MS with the real match sequences and find that, in both matches, the average momentum value during winning streaks is consistently above 0.5, while during losing streaks, it is below 0.5. This indicates the effectiveness of the MS metric in measuring consecutive wins and losses in tennis matches, as well as their efficient modeling of momentum at the point granularity. We also observe an intriguing trend: the momentum of the two players generally exhibits opposite trends, likely due to HydraNet's effective extraction of adversarial information. This adversarial relationship learning greatly enhances the MS's multi-granularity predictive capability for match progress.
Next, we analyze the influence of the momentum at the previous game or set on the current game's or set's outcome. We find that game-level prediction is 13\% stronger on average than set-level prediction, but both are less accurate than point-level prediction. This may be due to the shorter interval between games compared to sets, allowing the momentum from the previous game to more consciously influence the next.
Finally, we analyze the HM points. In w-1301, Carlos Alcaraz's HM point MS value is 0.6363, greater than 0.5, and he wins the match. We also find that he wins at this HM point. In w-1701, his MS value at the HM point is 0.4701, less than 0.5, and he loses the match, although he wins the point at that HM. Comparing these two matches and the match-level results in Table~\ref{tab:performance_metrics}, we infer that the MS value at the HM point is not solely determined by the outcome at the point level but is a composite measure of the player's performance during the first half of the match and the various influences they face.

\section{Conclusion}
In this paper, we introduce the MS metric to measure the momentum strength of players across different granularity levels in tennis tournaments. We also propose HydraNet, a momentum-driven SSD framework designed for MS construction. Additionally, we construct a large-scale cross-league tennis dataset, comprising millions of data points, for MS construction and HydraNet performance validation. Through extensive experimental analysis, we demonstrate HydraNet’s state-of-the-art performance in MS construction and validate the effectiveness of the MS metric for predicting match outcomes across various granularities. Our results reveal that momentum influences match outcomes to varying degrees across different granularity levels, and we also identify an intriguing ``half-court champagne'' phenomenon for the first time. We believe that the proposed MS metric, HydraNet framework, and the WID and USD datasets, along with the analysis of MS’s predictive ability in match progression, provide valuable insights and tools for future research on momentum in tennis and other sports. Moreover, these contributions also offer practical applications for tennis coaches and players in match analysis and strategy development.

\section{Acknowledgement}
This work has been supported by the National Natural Science Foundation of China (62302075, 62472062), the Innovation Support Program for Dalian High-Level Talents (2023RQ007), the Dalian Excellent Young Project (2022RY35), and the Dalian Science and technology Innovation Fund project (2024JJ12GX022).

\appendix
\section{Appendix}
\subsection{Details of Tennis Datasets}
\subsubsection{Data Cleaning}
We obtained player match information for the Wimbledon Championships (2012-2023) and the US Open (2013-2023) from the official tennis league websites using scripts, followed by data cleaning, deduplication, and format conversion to ensure data quality and consistency. The processed match data consists of a total of 1,028,340 entries. For all collected datasets, we first used scripts to select the necessary columns, sorting and removing any unnecessary ones. We then merged all the datasets to create the initial aggregate dataset. Next, we performed data cleaning and reconstruction for problematic or missing data. The key portion of the data cleaning and construction process is outlined below:
\begin{table*}[htbp]
\centering
\small 
\caption{Explanation of the factors in the WID and USD datasets.}
\resizebox{\textwidth}{!}{ 
\begin{tabular}{cllc}
\toprule
\textbf{No.} & \textbf{Variable} & \textbf{Explanation} & \textbf{Range} \\
\midrule
1 & match\_id & A unique identifier for each match, typically including details such as event and round. & N/A \\
2 & player1 & Full name of the first player (p1), denoting the participant in the match. & N/A \\
3 & player2 & Full name of the second player (p2), denoting the opponent in the match. & N/A \\
4 & elapsed\_time & Time elapsed from the start of the first point to the current point. & 0:00:00-23:59:59 \\
5 & set\_no & The ordinal number of the set within the match. & \{1, 2, 3...\} \\
6 & game\_no & The ordinal number of the game within the set, indicating the progression of the set. & \{1, 2, 3...\} \\
7 & point\_no & The ordinal number of the point within the current game. & \{1, 2, 3...\} \\
8 & p1\_sets\_won & Sets won by player1 throughout the match up to the current point. & \{0, 1\} \\
9 & p2\_sets\_won & Sets won by player2 throughout the match up to the current point. & \{0, 1\} \\
10 & p1\_games\_won & Games won by player1 within the current set. & \{0, 1\} \\
11 & p2\_games\_won & Games won by player2 within the current set. & \{0, 1\} \\
12 & p1\_score & Current score of player1 in the ongoing game. & \{0, 15, 30, 40\} \\
13 & p2\_score & Current score of player2 in the ongoing game. & \{0, 15, 30, 40\} \\
14 & p1\_serve & Indicator if player1 is serving. & \{0, 1\} \\
15 & p2\_serve & Indicator if player2 is serving. & \{0, 1\} \\
16 & points\_victor & The player who won the point. & \{1, 2\} \\
17 & p1\_points\_won & Points won by player1 in the current game. & \{0, 1\} \\
18 & p2\_points\_won & Points won by player2 in the current game. & \{0, 1\} \\
19 & p1\_points\_sum & Cumulative points won by player1 throughout the match. & \{1, 2, 3...\} \\
20 & p2\_points\_sum & Cumulative points won by player2 throughout the match. & \{1, 2, 3...\} \\
21 & game\_victor & The player who won the game. & \{1, 2\} \\
22 & set\_victor & The player who won the set. & \{1, 2\} \\
23 & p1\_ace & Aces served by player1, where the opponent could not return the serve. & \{0, 1\} \\
24 & p2\_ace & Aces served by player2, where the opponent could not return the serve. & \{0, 1\} \\
25 & p1\_winner & Winners struck by player1, where the opponent was unable to return the shot. & \{0, 1\} \\
26 & p2\_winner & Winners struck by player2, where the opponent was unable to return the shot. & \{0, 1\} \\
27 & p1\_double\_fault & Instances where player1 missed both serves and lost the point. & \{0, 1\} \\
28 & p2\_double\_fault & Instances where player2 missed both serves and lost the point. & \{0, 1\} \\
29 & p1\_unf\_err & Unforced errors by player1, indicating mistakes made without opponent pressure. & \{0, 1\} \\
30 & p2\_unf\_err & Unforced errors by player2, indicating mistakes made without opponent pressure. & \{0, 1\} \\
31 & p1\_net\_pt & Instances where player1 approached the net during the rally. & \{0, 1\} \\
32 & p2\_net\_pt & Instances where player2 approached the net during the rally. & \{0, 1\} \\
33 & p1\_net\_pt\_won & Points won by player1 while positioned at the net. & \{0, 1\} \\
34 & p2\_net\_pt\_won & Points won by player2 while positioned at the net. & \{0, 1\} \\
35 & p1\_break\_pt & Break point opportunities for player1 when the opponent is serving. & \{0, 1\} \\
36 & p2\_break\_pt & Break point opportunities for player2 when the opponent is serving. & \{0, 1\} \\
37 & p1\_break\_pt\_won & Break points converted by player1 when the opponent serves. & \{0, 1\} \\
38 & p2\_break\_pt\_won & Break points converted by player2 when the opponent serves. & \{0, 1\} \\
39 & p1\_break\_pt\_missed & Break points missed by player1 when the opponent serves. & \{0, 1\} \\
40 & p2\_break\_pt\_missed & Break points missed by player2 when the opponent serves. & \{0, 1\} \\
41 & p1\_distance\_run & Distance covered by player1 during the point, measured in meters. & \{0, 1\} \\
42 & p2\_distance\_run & Distance covered by player2 during the point, measured in meters. & \{0, 1\} \\
43 & p1\_points\_diff & Point difference between player1 and player2 in the match. & \{..., -1, 0, 1...\} \\
44 & p1\_game\_diff & Game difference between player1 and player2 in the match. & \{..., -1, 0, 1...\} \\
45 & p1\_set\_diff & Set difference between player1 and player2 in the match. & \{..., -1, 0, 1...\} \\
46 & p2\_points\_diff & Point difference between player2 and player1 in the match. & \{..., -1, 0, 1...\} \\
47 & p2\_game\_diff & Game difference between player2 and player1 in the match. & \{..., -1, 0, 1...\} \\
48 & p2\_set\_diff & Set difference between player2 and player1 in the match. & \{..., -1, 0, 1...\} \\
49 & p1\_serve\_speed & Average serve speed of player1, measured in miles per hour. & [0, 1] \\
50 & p2\_serve\_speed & Average serve speed of player2, measured in miles per hour. & [0, 1] \\
51 & p1\_serve\_depth & Average serve depth for player1, indicating how far the ball lands in the service box. & [0, 1] \\
52 & p2\_serve\_depth & Average serve depth for player2, indicating how far the ball lands in the service box. & [0, 1] \\
53 & p1\_return\_depth & Average return depth for player1, representing the distance from the baseline. & [0, 1] \\
54 & p2\_return\_depth & Average return depth for player1, representing the distance from the baseline. & [0, 1] \\
\bottomrule
\end{tabular}
}
\label{tab:3}
\end{table*}

\textbf{Return Depth:} Some matches had missing return depth data. We filled these missing values with 0. We also performed binary classification for known return depths, assigning 1 for ``D'' and 0 for ``ND''.

\textbf{Serve Depth:} Similarly, some matches had missing serve depth data. These missing values were also filled with 0. Binary classification was applied to known serve depths, assigning 1 for ``CTL'' and 0 for ``NCTL''.

\textbf{Current Game Score:} We removed all rows where the score was ``0X'' or ``0Y'' and rows where the server was listed as 0.

\textbf{Player's Game Score Difference, Set Score Difference, and Match Score Difference:} We calculated these differences by subtracting the corresponding statistics for the two players.

\textbf{Winning Point Per Round:} We assigned a value of 1 or 0 for each round, where 1 indicates the player won the point and 0 indicates they lost.

\textbf{Serve Speed:} To address the issue of missing server speed data, we performed reasonable imputation and applied a special maximum-minimum normalization to ensure the data is appropriately standardized for training and validation of HydraNet. First, we imputed missing serve speed data in two cases: (1) when the entire match’s serve speeds were missing, such as in the cases of Daniel Elahi Galan and Mikael Ymer in match 2023-wimbledon-1310, and Guido Pella and Roman Safiullin in match 2023-wimbledon-1311; (2) when sporadic missing serve speed data occurred. We observed that players like Roman Safiullin and Daniel Elahi Galan had complete serve speed data in other matches. For these players, we applied two imputation strategies: (1) For players with known serve speed data from other matches, we merged their known serve speed distribution with the global serve speed distribution of all players at a 1:1 ratio, and then imputed missing values based on this merged distribution. Additionally, considering the relationship between serve speed and performance in serving games, we applied a scoring mechanism that adjusted serve speeds based on game outcomes. (2) For players without known serve speed data from other matches and for sporadic missing data, we used the global serve speed distribution to impute missing values and applied the same scoring adjustment based on game outcomes. After completing the imputation, we referred to relevant literature on professional tennis data standards, where a serve speed of >124 mph is considered good, 96 mph < serve speed \( \leq \) 124 mph is considered average, and serve speed < 96 mph is considered below average. To better reflect the influence of serve speed on performance scores, we applied a special maximum-minimum normalization formula to the serve speed data:
\begin{equation}
    Y = 2 \times \left( \frac{X - X_{\min}}{X_{\max} - X_{\min}} \right) - 1
\end{equation}
where \(Y\) is the normalized serve speed value, ranging from [-1,1], and \(X\) is the original serve speed. Upon reviewing the normalized values, we found that serve speeds below 94 mph were transformed into values below 0.5, speeds below 105 mph became values below 0, speeds above 105 mph were converted into values above 0, and speeds above 120 mph were transformed into values above 0.5. These transformations align with our research findings and reflect a continuous scale that better captures the differences in serve speeds.

\textbf{Running Distance:} To stabilize model training, we applied Z-Score normalization to the running distance data:
\begin{equation}
z_i = \frac{x_i - \mu}{\sigma}
\end{equation}
where \( z_i \) represents the normalized running distance data point, \( x_i \) is the original data point, \( \mu \) denotes the mean of the data, and \( \sigma \) denotes the standard deviation of the data.
\subsubsection{Data Format}
Table~\ref{tab:3} provides detailed information about the WID and USD datasets we constructed, encompassing 54 factors. In this study, we selected 32 factors strongly correlated with Serve, Return, Psychology, and Fatigue (as shown in Table~\ref{tab:1}) for the direct construction of the MS. Additionally, 11 other factors, such as match\_id, player1, player2, set\_no, game\_no, point\_no, p1\_points\_won, p1\_games\_won, p2\_games\_won, p1\_sets\_won and p2\_sets\_won , were indirectly employed for the MS parameter analysis.
\vspace{-5pt}
\subsection{Details of Core Parameters in MSSD}
For the construction of the four core parameters in MSSD learning, We begin by applying a linear transformation and dimensionality adjustment to the input feature matrix \(\mathbf{G} \in \mathbb{R}^{(1+L) \times D_{\text{in}}}\), where \(D_{\text{in}} = 2 \times D_{\text{inner}} + 2 \times D_{\text{state}} + H\). The transformed features are then projected into three subspaces:
\begin{equation}
z, \mathbf{x}_{\mathrm{BC}}, dt = \text{split}(\boldsymbol{\omega} \cdot \mathbf{G} + \mathbf{b}),
\end{equation}
where \(\boldsymbol{\omega}\) is the weight matrix for the linear transformation, \(\mathbf{b}\) is the bias term, and \(\text{split}(\cdot)\) divides the resulting matrix into:
\( z \in \mathbb{R}^{(1+L) \times D_{\text{inner}}} \), used for residual normalization; 
\( \mathbf{x}_{\mathrm{BC}} \in \mathbb{R}^{(1+L) \times (D_{\text{inner}} + 2 \times D_{\text{state}})} \), which captures dynamic temporal interactions; and 
\( dt \in \mathbb{R}^{(1+L) \times H} \), representing temporal weighting factors for attention heads. Here, \(H\) denotes the number of attention heads, \(P\) is the dimension of each attention head, \(D_{\text{inner}} = H \cdot P\), and \(D_{\text{state}} = d\).

Building on the construction of initial temporal weighting factors \( \mathbf{dt} \), we now compute the refined temporal decay matrix \( \mathbf{A} \), which integrates both the temporal decay factor and dynamic adjustments. The computation begins by refining the dynamic weight \( dt \) using the Softplus activation function with a learnable bias \( b_{dt} \), followed by combining it with the base temporal decay factor. The final formulation is defined as:
\begin{equation}
{\mathbf{A}} = -\exp(\mathbf{A}_{\text{log}}) \cdot \mathrm{Softplus}(dt + b_{dt}),
\end{equation}
where \(\mathbf{A}_{\text{log}} \in \mathbb{R}^H\) is a learnable vector that governs the base temporal decay factor. \(\exp(\cdot)\) denotes the element-wise exponential function. \( b_{dt} \in \mathbb{R}^H \) is a learnable bias vector. \(\mathrm{Softplus}(\cdot)\) denotes a smooth approximation of the ReLU function. The resulting matrix \( {\mathbf{A}} \in \mathbb{R}^{(1+L) \times H} \) effectively combines these components, enabling precise modeling of temporal dynamics across attention heads.

From dynamic temporal interactions \(\mathbf{x}_{\mathrm{BC}}\), we further derive the key parameters \( x \), \( \mathbf{B} \), and \( \mathbf{C} \) by applying the \(\text{split}(\cdot)\) function. The core dynamic feature matrix \( x \in \mathbb{R}^{(1+L) \times D_{\text{state}}} \) is used for subsequent computations, while the matrix \( \mathbf{B} \in \mathbb{R}^{(1+L) \times D_{\text{state}}}\) and matrix \( \mathbf{C} \in \mathbb{R}^{(1+L) \times D_{\text{state}}}\) are essential for modeling state-space transformations. To ensure compatibility with MSSD learning, the feature matrix \( x \) is reorganized to fit the multi-head attention mechanism, resulting in a reshaped dimension of \(\mathbb{R}^{(1+L) \times H \times P}\). Similarly, the mapping matrices \( \mathbf{B} \) and \( \mathbf{C} \) are further expanded to \(\mathbb{R}^{(1+L) \times 1 \times D_{\text{state}}}\), enabling them to align with the state-space transformations required in the MSSD framework. Thus, the core parameters for MSSD learning have been fully constructed.
\bibliographystyle{unsrt}
\biboptions{sort&compress}

\bibliography{reference}


\end{document}